\documentclass[letterpaper]{article}
\usepackage[preprint]{neurips_2019}
\usepackage{times}
\usepackage{helvet}
\usepackage{courier}
\usepackage{lipsum}
\usepackage{amsmath}
\usepackage{amssymb}
\usepackage{graphicx}
\usepackage{bigstrut}

\begin{document}

\title{ Graph-Hist: Graph Classification from Latent Feature Histograms \\
With Application to Bot Detection}
\author{Thomas Magelinski, David Beskow, Kathleen M. Carley\\
CASOS, School of Computer Science, Carnegie Mellon University\\
5000 Forbes Ave. Pittsburgh, PA, 15213
}
\maketitle
\begin{abstract}
    Neural networks are increasingly used for graph classification in a variety of contexts.
    Social media is a critical application area in this space, however the characteristics of social media graphs differ from those seen in most popular benchmark datasets.
    Social networks tend to be large and sparse, while benchmarks are small and dense.
    Classically, large and sparse networks are analyzed by studying the distribution of local properties.
    Inspired by this, we introduce Graph-Hist: an end-to-end architecture that extracts a graph's latent local features, bins nodes together along 1-D cross sections of the feature space, and classifies the graph based on this multi-channel histogram.
    We show that Graph-Hist improves state of the art performance on true social media benchmark datasets, while still performing well on other benchmarks.
    Finally, we demonstrate Graph-Hist's performance by conducting bot detection in social media. 
    While sophisticated bot and cyborg accounts increasingly evade traditional detection methods, they leave artificial artifacts in their conversational graph that are detected through graph classification.
    We apply Graph-Hist to classify these conversational graphs. In the process, we confirm that social media graphs are different than most baselines and that Graph-Hist outperforms existing bot-detection models.
\end{abstract}

\section{Introduction}
Given the success of traditional machine learning, interest in geometric learning has grown in recent years. 
Geometric learning seeks to extend machine learning models beyond euclidean data to include objects such as graphs, point clouds, and manifolds.
Non-euclidean data structures are information-rich, as they can describe data that traditional structures cannot.
For example, a traditional data structure may contain attributes of a group of individuals, while a graph or network can also encode the \textit{relationships} between the individuals.
Thus, new algorithms to leverage this type of information can bring new insights.

For graphs in particular there are three main problems: node classification, link prediction, and graph classification.
Here, we focus on the last task, graph classification.
In this problem, each input sample is a graph, which has a corresponding category or label.
The goal is to create a model that takes an entire graph as an input, and assigns it to the correct class.

Graph classification is gaining interest in part due to the variety of domains it may be applied to. 
The same models that can classify proteins based on their structure may also be used to classify social media conversations.
We identify a new application which is highly relevant in today's socio-political landscape: bot classification on social media.
Automated accounts called bots are increasingly used in online information operations to manipulate both networks (virtual social links) and the narratives that transit these networks.
Since bots operate \textit{through} networks, their network structure can be used to identify them.

While many problems regarding social media data can be posed under a graph classification framework, few prior models focus on this domain. 
Much of the prior work in graph classification focuses on benchmark data that does not reflect the typical structure of social media data. 
Specifically, social media graphs or networks tend to have large nodesets and low density, while benchmark datasets tend to have less than 100 nodes and are quite dense. 

In this work, we develop a new graph classification architecture inspired by classical network analysis.
In analysis of large networks, it is common practice to calculate local (node-level) features, and study the distribution.
Here, we use an end-to-end graph-convolutional architecture to extract local latent features and classify the graph based on the distribution of these features. 
Due to the high dimensionality of the feature space, we instead use 1-D cross sections of the distribution in the form of a multi-channel histogram.
Since this procedure classifies graphs based on histograms of latent features, it has been named Graph-Hist.

In the following sections, we review prior work in graph classification, explain our architecture, demonstrate Graph-Hist's ability to achieve state of the art results on social media benchmarks, and finally demonstrate real-world application of our model through a case study of bot classification on Twitter data. 

The field of graph learning has expanded rapidly since the notation for Graph Neural Networks (GNNs) were first introduced by Gori et al. (\cite{gori2005new}). The work from then to 2018 is well summarized by Wu et. al (\cite{wu_comprehensive_2019}).

GNNs for graph classification are typically based on a type of \textit{graph convolution}. 
Traditional convolutional networks have proved extremely successful at learning shape features in the euclidean domain, such as image classification, however translating this operation to the graph domain is difficult due to irregularities in graph structure (\cite{krizhevsky2012imagenet}).
Graph convolutions usually fall into one of two approaches: spectral or spatial.
Spectral methods stem from efforts to extend traditional signal processing techniques to graph signals, or Graph Signal Processing (\cite{shuman2013emerging}).
Spectral approaches typically use the symmetric normalized Laplacian, shown in Equation \ref{eq:laplacian}. 
Many spectral-based approaches like ChebNet relied on eigenvalue calculations, making them computationally costly (\cite{defferrard2016convolutional}). Spatial methods on the other hand, operate on the local structure of the graph. In spatial methods such as GraphSage, nodes aggregate information from their neighbors (\cite{hamilton2017inductive}). 

Kipf and Welling introduced a model that bridged the gap between the two: it is an approximation of a spectral convolution, but it is localized in space (\cite{kipf2016semi}).
This model uses the propagation rule shown in Equation \ref{eq:kipf}, where trainable weights, $W$, are multiplied into learned node features, $Z^l$, and then into the Laplacian $L$. The process starts by using given node features, $X$, as an input: $Z^0 = X$.
\begin{equation}
    \label{eq:kipf}
    Z^{(l+1)} = \sigma(LZ^lW)
\end{equation}
Many architectures for graph classification now build upon this convolutional structure, including two works we draw from here: Sortpool and Capsule Graph Networks (\cite{zhang_end--end_2018,xinyi_capsule_2019}).

Zhang et al, replaces the normalized Laplacian with the random-walk Laplacian, and draws parallels to the the Weisfeiler-Lehman subtree kernel (\cite{shervashidze2011weisfeiler}).
This effectively gives node embeddings, which they then sort and either truncate or pad to a fixed size, hence the name SortPool.
While the sorting procedure gives some spatial relationship to the nodes, the truncating/padding procedure either drops important information, or adds erroneous data when datasets have high variance in graph size, which is often the case in social media datasets. 
Xinyi and Chen have also used this GNN structure, but applied attention to handle the differences in graph size (\cite{xinyi_capsule_2019}).

Tixier et al. take a different approach (\cite{tixier_graph_2017}). 
They first assume that node embeddings are given. Node embeddings can be obtained in a number of unsupervised ways, most of which attempt to preserve the network-based distance between nodes in the embedded space through operations like random walks (\cite{perozzi2014deepwalk}).
They then compress this high-dimensional embedding into a multi-channel image by looking at cross sections of consecutive principle components from principle component analysis (PCA). 
Finally, they use a standard image classifier architecture to classify the graphs.
This approach achieved good results, but has two shortcomings.
First, the lack of end-to-end architecture results in embeddings that may work well for spatial preservation, but poorly for graph classification.
Second, their pairing of PCA dimensions is somewhat arbitrary.

Here, we effectively combine and apply a powerful CNN architecture like that used by Tixier et al to the expressive node embeddings from GNNs, as in Kipf, Zhang, and Xinyi.
The previously missing piece to attaching these two methods is a differentiable operation that converts node embeddings into a format that CNN can leverage.
To define such an operation we draw from classical network science.
Networks are typically analyzed by studying the distribution of their local features (\cite{wasserman1994social}).
The most common of such analyses is performed on the degree distribution, which has been used to classify networks of different types, such as scale-free or small world network.
This individual analysis of feature histrograms inspired the binning mechanism introduced here.
Our binning operation approximates the full node embedding distribution into a multi-channel histogram, which is easily inputted to a standard CNN.

\section{Graph-Hist}
In graph classification a set of labeled graphs is given as $\{(G_1, y_1), (G_2, y_2), ...\}$. There are $m$ potential labels, so $y_i \in [0, 1, ...m]$ for all $i$.
Our task is to find a mapping from a graph to its label.
That is, to find a mapping $f$ such that $f: G_i \rightarrow y_i$.
In practice, we train a model to minimize the cross entropy loss for this classification: $\ell =  - \sum_i \sum_{c=1}^m \mathcal{I}(y_i == c) \log(p_{i,c})$, where $\mathcal{I}$, is an indicator function outputting 1 if given a true statement, and 0 otherwise and where $p_{i,c}$ is the probability the model assigns training example $i$ to belong to class $c$.

For each graph, we will operate on the adjacency matrix, $A$. 
Elements in $A$ indicate the links in $G$ such that $A_{i,j}=w$, if nodes $i$ and $j$ are linked with weight $w$, and $A_{i,j}=0$ otherwise. 
We also assume the graph has self loops, $A_{i,i}=1$.
Additionally, each node in graph $G$ has $f$ features, encoded in $X \in \mathbb{R}^{n \times f}$.
In most example datasets, no node features are given. In these cases we use the the node degree and an identity feature. 
The degree of node $i$ is given by $d_i = \sum_j A_{i,j}$. From there, the degree matrix is constructed as $D = \text{diag}(d)$

\begin{figure*}
    \centering
    \includegraphics[width=\linewidth]{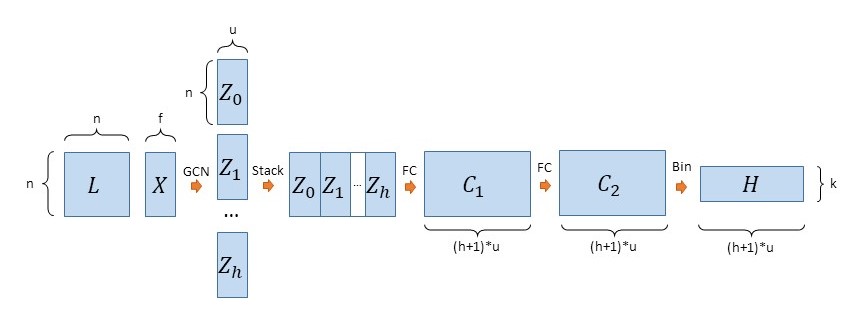}
    \caption{The general architecture used in this work. The output, $H$, is the multi-channel histogram that is then classified using a 1-dimensional variant of Lenet-5, as described in the Histogram Classification section. }
    \label{fig:model}
\end{figure*}

Given this setting, a diagram of our proposed architecture is given in Figure \ref{fig:model}. 

\subsection{Graph Convolution}
Our variant of graph convolution is given in Equation \ref{eq:GCN}, where $W_s \in \mathbb{R}^{f \times u}, b_s \in \mathbb{R}^{u},$ are trainable weight and bias parameters, respectively, and where $\sigma$ is a non-linear activation function.
This GCN relies on the normalized Laplacian matrix (shown in Equation \ref{eq:laplacian}), making it similar to \textit{spectral} GCNs.
Spectral GCNs were popularized by Kipf and Welling when they provided a local approximation to a ChebNet, greatly reducing the computational cost (\cite{kipf2016semi}). 
\begin{align}
    Z_s &= \sigma(\bar{L}^sXW+b) \label{eq:GCN} \\
    \bar{L} &= D^{-\frac{1}{2}}(D-A)D^{-\frac{1}{2}} \label{eq:laplacian}
\end{align}
These spectral GCNs serve as a local approximation to the more general convolutional framework that was originally proposed by Bruna et al. and have strong underpinnings in graph signal processing (\cite{bruna2013spectral,shuman2013emerging}).
Another possibility would have been the random walk Laplacian, $L^{rw} = I - D^{-1}A$, which encodes the probability of transitioning from node to node. 
The random walk Laplacian has been used in a variety of works, both for graph classification and node embedding (\cite{scarselli_graph_2009,zhang_end--end_2018,hamilton2017inductive}).

However, given the underpinnings in graph signal processing and  the success of recent spectral models, we move forward with this definition.

Previous graph classification architectures like Sortpool and Capsule Graph stack GCNs such that embeddings from the first GCN are the new input features to the second GCN, which then outputs features for the third, and so forth.
This form of stacking allows the features to be passed beyond direct neighbors. A $h$-level stacking allows nodes to aggregate features within a its neighborhood of radius $h$.

Additionally, Zhang et al. show that stacked GCNs provide a continuous analog to the graph coloring problem and the Weisfeiler-Lehman subtree kernel (\cite{zhang_end--end_2018,shervashidze2011weisfeiler}).

Despite this, GCN stacking necessitates that feature aggregation from a node's extended neighborhood is reliant on aggregation closer to the source node.
In our framework, aggregation is obtained independently (through the power of the Laplacian matrix).
While previous works do not explicitly use powers of the Laplacian, they do so implicitly through the stacking of GCN modules.
Experiments with our approach show slight improvements in performance while allowing for parallelization if memory allows.
Additionally, independence of GCNs allows for the use of a bias term, which is not natural in stacked-GCN architectures, since they would be multiplied back into the Laplacian at the next level.
Lastly, we include $Z_0$ in our GCN step, which reduces the Laplacian to the identity matrix, thus providing a standard fully connected sub-module from the node features.

\subsection{Fully Connected Combination Layers}
As in Sortpool, Graph Capsules, and others, the node embeddings obtained from $Z_0, ...Z_h$ are concatenated to give node embeddings of dimension $(h+1)*u$. 
In the succeeding step, the intra-dimension properties are lost (more details on why in the following section). 

To minimize the information loss, the embeddings are passed through two fully connected layers, which can capture these nonlinear intra-dimension properties.  
For simplicity we've chosen layers that have the same dimension as the initial embeddings, $(w+1)*h$.
This step is similar to the initial convolution applied to the final embeddings in Sortpool, with differences being in the size of the output, and the activation function.

\subsection{Node Binning}
One of the fundamental challenges in graph classification is that each input graph can have a different number of nodes, $n$, in its nodeset.
Thus, graph classification algorithms which rely on node embeddings must find some way of transitioning from a variable size $n$ to a fixed size $k$. 

Sortpool, for example, re-shapes the data by setting a threshold $k$. After sorting, the top $k$ nodes are selected, and the rest are dropped. If the graph has more than $k$ nodes, zeros are appended to the nodeset until it is size $k$.
As previously discussed, this is sub-optimal from a data preservation point of view.

We solve this problem through a binning procedure.
The input space is discretized, and the number of nodes falling into each discrete bin is counted.
Then, a standard convolutional architecture can be applied to the obtained density function.
However, effective node embeddings are typically high in dimension, making discretization over the full space intractable.
The 2D-CNN approach taken by Tixier et al.  approximates the distribution using principle components (\cite{tixier_graph_2017}).
They take 2-D cross sections of the ascending principle components, stacking them as channels of an image.
Given the image output, a standard image classifier can then be applied.
While this achieved good results, it relies on pre-processing (using given node embeddings and calculating the principle components) and cannot update node embeddings to improve performance.

Here we provide a differentiable alternative that does not rely on a dimension-pairing scheme.
Instead, we bin the data along one-dimensional cross sections of each dimension, resulting in a histogram with $(h+1)*u$ channels.
The number of bins is a tuneable parameter, $k$.

The derivative of the loss function, $\ell$, can then be propagated through the binning layer by average weighted of the bin gradients, as shown in Equation \ref{eq:bindir2}. First, the distance $d$ to the bin centers, $b$, are calculated, making $d\in \mathbb{R}^{k \times n \times (h+1)*u}$. Then, weighted average of the bin gradients are taken, allowing bins closer to nodes to have more pull than bins further away. Thus, each bin pulls nodes towards it if its gradient is positive, and pushes nodes away if its gradient is negative. The amount of pull is proportionate to its distance from the nodes and is controlled by $\alpha$, which we have set to $\alpha=20$. While the activation function for $C_2$ does not necessarily have to be $tanh$, it must be bounded. Without a bounded activation function bin boundaries could not be predefined to capture all of the nodes placements. With $tanh$, for example, all output values will be between -1, and 1, so if $k=10$, evenly spaced non-overlapping bins will be of length $0.2$, and will capture all potential node placements.

\begin{align}
    d & = b - C_2 \label{eq:bindir1} \\
    \frac{\partial \ell }{\partial C_{2,l,j}} &= \frac{1}{ \sum_{i=1}^k e^{-\alpha |d_{i,l,j}|}} \sum_{i=1}^k e^{-\alpha |d_{i,l,j}|}\text{sign}(d_{i,l,j}) \frac{\partial \ell }{\partial H_{i,j}} \label{eq:bindir2}
\end{align}

Again, this process does lose the co-variance relationship between the dimensions of the distribution.
However, the combination layers overcome this simplification since we are using an end-to-end architecture.
Nodes will be pushed along these 1-D cross sections during back-propagation such that a classification can be made.
The effectiveness of this approach is demonstrated on standard benchmark data in the Experiments section and in a case study for a new application, bot classification, in the Case Study section.

\subsection{Histogram Classification}
\label{sec:lenet}
Finally, the multi-channel histogram can be classified using a traditional convolutional architecture.
Tixier et al used a variant of LeNet-5 to classify graphs based on on 2-D cross sections of predefined node embeddings in (\cite{tixier_graph_2017}).
The CNN achieved 99.45\% accuracy on the MNIST handwritten digit classification task. 
We slightly modify the architecture to suit the 1-dimensional data that we have obtained from the previous steps.

As in \cite{tixier_graph_2017}, $H$ is passed to 4 sub-modules, with filter sizes of $f=3, 4, 5, \text{and } 6$, respectively.
A sub-module is performed as follows. The input data is convolved over with its filter size $f$ and a stride of 1, to 64 output channels. 
Then, max pooling is performed with size and stride 2. 
The convolution is performed again, but with 96 output channels. 
Simultaneously, $H$ is passed to a convolution with $f=(h+1)*u$, thus capturing the entire histogram with 96 output channels.
The sub-module outputs and the full-histogram convolutions are concatenated and connected to a fully connected layer of size 256.
Lastly, the 256 unit layer is connected to a softmax output that classifies the graph. Dropout layers were placed before all fully connected layers in the histogram classifier.
The activation function used was ReLU.
The three changes from the original classifier given are: the whole-histogram convolution is added, the 128 hidden unit layer was changed to 256 units, and the model was adapted to its 1-dimensional analog.

The entire model, then, can be trained in an end-to-end manner.

\section{Experiments}\label{sec:experiments}
\subsection{Benchmark Datasets and Methods}
There are many potential benchmark datasets for graph classification, however few of them are social networks, and even fewer resemble the type of networks seen in real world social media data. 
Real world social media networks are typically large and sparse (\cite{onnela2007structure}).

However, most benchmark datasets are relatively dense and have nodesets with less than 100 nodes. 
With this in mind, we have selected 6 popular benchmark datasets, displayed in Table \ref{tab:benchmark_data_info}.
The datasets have been obtained from Kersting et al's collection, but were created by Yanardag and Vishwanathan (\cite{all_datasets,yanardag2015deep}).

\begin{table*}
    \centering
    \caption{A summary of the datasets studied. Numbers for nodes and edges are averages.}
    \resizebox{\columnwidth}{!}{%
    \begin{tabular}{l|c|c|c|c|c|c|c}
        \hline
        \hline
        \textbf{Dataset} & IMDB-B & IMDB-M & COLLAB & REDDIT-B & REDDIT-5K & REDDIT-12K & Bots   \\
        \hline
        Graphs & 1000 & 1500 & 5000 & 2000 & 4999 & 11929 & 14962\\
        Classes & 2 & 3 & 3 & 2 & 5 & 11 & 2\\
        Nodes & 19.77 & 13.00 & 74.49 & 429.63 & 508.52 & 391.41 & 7294\\
        Edges & 96.53 & 65.94 & 2457.78 & 497.75 & 594.87 & 456.89 & 11034\\
        \hline
    \end{tabular}
    }
    \label{tab:benchmark_data_info}
\end{table*}

The IMDB datasets are movie collaboration datasets. Nodes are actors/actresses and links represent co-appearance in a movie. The graphs are ego networks, and the task is to classify the genre that an ego network belongs to. This dataset is somewhat challenging because movies may belong to more than one genre, but may only be given one label.

COLLAB was derived from scientific collaboration data in three fields: High Energy Physics, Condensed Matter Physics, and Astro Physics.
Each graph is an author's ego network, and the task is to identify which field they work in.

All three of the Reddit datasets were scraped from the social media platform Reddit, using their API.
Nodes in the graph are Reddit users, and links are created by direct replies in the discussion.
In the binary dataset, the graphs either come from question-and-answer subreddits, or discussion-based subreddits. The task is to identify which type of subreddit the conversational graph comes from.
In the 5k and 12k, datasets, the task is to identify the specific subreddit that the graph belongs to.

We place greater emphasis on the Reddit datasets, as they are the only social media classification tasks. Table \ref{tab:benchmark_data_info}, illustrates the importance of this distinction. The graphs in the Reddit datasets tend to be an order of magnitude bigger in terms of nodeset size, and two orders of magnitude lower in density.

We have selected 5 different methods to compare our results against, namely Anonymous Walk Embeddings, Sortpool, DiffPool, CapsGNN, and 2D CNN, (\cite{ivanov2018anonymous,zhang_end--end_2018,ying2018hierarchical,tixier_graph_2017}). 
These methods were selected to reflect state-of-the-art classification results, and to compare our results against methods from which we have built upon.
To the best of our knowledge, the current state-of-the-art performances are shown for every dataset.
The accuracies and standard deviations are reported in Table \ref{tab:benchmark_data_results} based on the values reported in initial publication.
Because of this, not every dataset has a value for every method.
Fey and Lenssen have introduced Pytorch Geometric, a library with implementations of many geometric learning algorithms (\cite{fey2019fast}).
Some gaps are filled by using values reported from their implementations.
Anonymous Walk Embeddings is the only kernel approach compared against, so it is separated in Table \ref{tab:benchmark_data_results}.

\subsection{Experimental Setting}
The general architecture used for all experiments is illustrated in Figure \ref{fig:model}.
Graph-Hist was implemented in Pytorch. 
The hyperbolic tangent function was used for all activation functions leading up to LeNet.
We used the ReduceLROnPlateau scheduler with an initial learning rate of $\alpha=1e-4$, a factor of 0.5, a patience of 2, a cooldown of 0, and a minimum learning rate of $1e-7$. 
We used stochastic gradient descent with a mini-batch size of 32.
We terminated training after 9 consecutive epochs without progress in the testing loss. 

We then tuned parameters to each dataset in the search space $n \in [2, 4, 6, 8]$, $hs \in [32, 64, 128, 256]$, $d \in [0.2, 0.8]$. Parameters were selected based on their performance on the test set.
The final parameters for each dataset is given in Table \ref{tab:model_parameters}.

Finally, we performed 10-fold cross-validation on each of the datasets using the parameters in Table \ref{tab:model_parameters}. The mean accuracy and its standard deviation is reported for each dataset in Table \ref{tab:benchmark_data_results}. Graph-Hist advances state-of-the-art classification in all 3 of the social media benchmarks. It also beats state-of-the-art results for IMDB-B, and obtains second place results for the remaining two datasets.

We recognize that there are many more hyperparameters that could be tuned, like the batch size, and that even the size of transformations like $C_1$ and $C_2$ could be tuned.
Exploring these possibilities is left for future work, but could result in even better results than those demonstrated here.

\begin{table}
    \centering
    \caption{The parameters used on each dataset.}
    \begin{tabular}{l|c|c|c|c}
        \hline
        \hline
        \textbf{Dataset} & k  & h &  u  & d \\
        \hline
        IMDB- B          & 50 & 2 & 128 & 0.8 \\
        IMDB-M           & 25 & 4 & 128 & 0.8 \\
        COLLAB           & 25 & 2 & 256 & 0.2 \\
        REDDIT-B         & 25 & 6 & 64  & 0.8 \\
        REDDIT-5K        & 25 & 8 & 64  & 0.8 \\
        REDDIT-12K       & 25 & 2 & 64  & 0.8 \\
        Twitter Bots     & 25 & 2 & 8   & 0.5 \\
        \hline
    \end{tabular}
    \label{tab:model_parameters}
\end{table}

\begin{table*}
    \centering
    \caption{Benchmark dataset accuracies from 10-fold cross-validation. Top-2 scores are emboldened, the state-of-the-art score is marked with an asterisk. Scores are shown as reported in publication, so not all datasets are represented or are shown with their standard deviation for every method.}
    \resizebox{\columnwidth}{!}{%
    \begin{tabular}{l|c|c|c|c|c|c}
        \hline
        \hline
        \textbf{Dataset} & IMDB-B & IMDB-M & COLLAB & REDDIT-B & REDDIT-5K & REDDIT-12K \bigstrut \\
        \hline
        AWE        & $\mathbf{74.5 \pm 5.8}$ & $\mathbf{51.5 \pm 3.6}^*$ & $73.9 \pm 1.9$ & $87.9 \pm 2.5$ & $50.5 \pm 1.9$ & $39.2 \pm 2.1$ \\
        \hline
        Sortpool   & $72.4 \pm 3.8$ & $47.8 \pm 0.8$ & $77.7 \pm 3.1$ & $74.9 \pm 6.7$ & - & -  \\
        DiffPool   & $72.6 \pm 3.9$ & - & $78.9 \pm 2.3$ & $\mathbf{92.1 \pm 2.6}$ & - & 47.1  \\
        CapsGNN    & $73.1 \pm 4.8$ & $\mathbf{50.3 \pm 2.6}$ & $\mathbf{79.6 \pm 0.9}^*$ & - & $\mathbf{52.9 \pm 1.5}$ & $46.6 \pm 1.9$ \\
        2D CNN     & $70.4 \pm 3.8$ & - & $71.3 \pm 2.0$ & $89.1 \pm 1.7$ & $52.1 \pm 2.2$ & $\mathbf{48.1 \pm 1.5}$ \\
        \hline
        Graph-Hist & $\mathbf{74.7 \pm 3.9}^*$ & $\mathbf{50.3 \pm 3.6}$ & $\mathbf{79.2 \pm 2.0}$ & $\mathbf{92.2 \pm 2.2}^*$ & $\mathbf{55.0 \pm 1.7}^*$ & $\mathbf{49.2 \pm 1.0}^*$  \bigstrut \\ 
        \hline
    \end{tabular}
    }
    \label{tab:benchmark_data_results}
\end{table*}

\begin{table}
    \centering
    \caption{Bot detection F1 Precision and Recall scores. All models but Botometer trained on debot data. Top-2 F1 scores are emboldened, the state-of-the-art score is marked with an asterisk.}
    \begin{tabular}{lccc}
        \hline
        \hline
        \textbf{Model} & F1 & Precision & Recall    \\
        \hline
        Botometer        & 0.524 & 0.858 & 0.377\\
        Debot            & 0.012  & 1.00 & 0.006\\
        bot-hunter Tier1 & 0.656  & 0.821 & 0.546\\
        bot-hunter Tier2 & \textbf{0.687}  & 0.691 & 0.683\\
        bot-hunter Tier3 & 0.599  & 0.837 & 0.466\\
        \hline
        Graph-Hist       & $\mathbf{0.740}^*$ &  0.683 & 0.807 \\
        \hline
    \end{tabular}
    \label{tab:major_test}
\end{table}

\section{Case Study}\label{sec:case_study}
\noindent Automated accounts called bots are increasingly used in online information operations to manipulate both networks (virtual social links) and the narratives that transit these networks.  In doing so, state and non-state actors can artificially manipulate the online marketplace of belief and ideas.  To battle this rise in bots, researchers at industry, government, and academia have developed increasingly sophisticated algorithms to detect these nefarious accounts.  These research efforts have led to a ``cat and mouse'' cycle in which increasingly sophisticated algorithms are required to detect increasingly sophisticated automated accounts.  Early detection models identified tell-tale indicators of automated activity such as stolen identities, lack of normal human circadian rhythms, anonymous attributes (lack of profile picture, random string screen name, etc), and a low follower/followee ratio.  These features, however, are relatively easy for a bot ``puppet-master'' to manipulate in order to remain undetected.  

It is much harder for these same bot ``puppet-masters'' to change the artificial features of the social and communication networks that they inhabit.  These social and communication networks (following, retweeting, mentioning, replying) lack the overlapping social integration of human social and communication links. Thus, we exploit the structure of these communication networks directly using Graph-Hist. We find that this approach generalizes to new datasets better than current alternative approaches.

\subsection{Building Networks}
We built the conversational network that a Twitter account inhabits in the same manner as (\cite{besow2018asonam}).  This approach combines the timelines of the target account and their followers to build the larger conversation.  This method was selected because it creates a comprehensive ego network while overcoming API rate limiting constraints and expediting the time it would take to collect the data (target collection is 5 minutes per account). While 5 minutes per account seems long, this is trivial compared to the hours or days that it would take to build a single ego network based on friends/followers connections.  The properties of these networks are summarized in Table \ref{tab:benchmark_data_info}.

Again, the differences between social media networks and standard benchmarks are pronounced. The Twitter networks are 2 orders of magnitude larger than the non-social media benchmarks in terms of nodeset size. The Twitter network densities are also 3 orders of magnitude smaller than those of the standard benchmarks.

\subsection{Previous Work in Bot Detection}

For the past decade, increasing numbers of researchers have worked on developing algorithms to detect increasingly sophisticated bots.  These models can be broadly separated into supervised machine learning models, unsupervised models, and graph based models.  These in turn have also create several prominent tools that are used in social cyber security workflows, including the Botometer (\cite{davis2016botornot}), Bot-hunter (\cite{beskow2018botHunter}), Debot (\cite{chavoshi2016debot}), and Botwalk (\cite{minnich2017botwalk}) algorithms.

Most of the graph and community detection methods have been conducted on Facebook, where these bots are at times called Sybils.  These include random walk approaches like Sybil-Guard (\cite{yu2006sybilguard}), Sybil-Resist (\cite{ma2014sybil}) and Sybil-Rank (\cite{cao2012aiding}).  Other models relax some of the assumptions and use trust propagation approaches such as the Sybil-Fence method (\cite{cao2013sybilfence}).

Supervised models include traditional machine learning with SVM (\cite{lee2014early}), Na\"{i}ve Bayes (\cite{chen2014feature}), and Random Forest (\cite{ferrara2016rise}) models trained on features extracted Twitter tweet objects and user objects.  Other methods have attempted to classify accounts based only on their text (\cite{kudugunta2018deep}) or their screen name (\cite{beskow2018random}).  Several of the available models like Botometer (\cite{davis2016botornot}) and Bot-hunter (\cite{beskow2018botHunter}) are classic supervised machine learning models.

Several unsupervised methods have also emerged, largely focused on identifying underlying patterns produced by certain types of bots.  These include clustering algorithms (\cite{benigni2017online}) and anomaly detection algorithms like the BotWalk algorithm (\cite{minnich2017botwalk}).

Most of these models leverage account data and account history while graph based models are focused finding patterns in the conversation and connections.  Not many models focus on the larger conversational ego-network surrounding the account.  Only one supervised machine learning model has attempted to bring network science metrics (centrality, simmelian ties, triadic census, etc) from these ego networks into their feature space (\cite{besow2018asonam}).  Rather than using network metrics as proxies for the network itself, we approached this same problem with geometric learning over the entire graph.  

\subsection{Bot Classification Results}

For the case study, we built training data of bot accounts that have been labeled by the Debot unsupervised algorithm.  The Debot algorithm uses warped correlation to identify ``correlated'' Twitter bots (bots that post the same content at roughly the same time) (\cite{chavoshi2016debot}).  Debot has demonstrated high precision identifying this special class of bots, has been used to train classic supervised bot detection models with strong results, and thus was used to label bot data for training.  Non-bot ``human'' data was randomly sampled from the Twitter 1\% Stream.  Our training data consisted of 8,842 bots and 6,120 human accounts and their associated conversational networks.  

We developed a separate test dataset to compare against other state of the art algorithms as well as measure generalizabilty.  The final test data was created by manually annotating 337 bot accounts focused on propaganda and other manipulation.  Emphasis was made to ensure this test data did not overlap with any training data used by the models tested.  The test dataset was balanced with 337 bot accounts and 337 human accounts.  

For an evaluation metric we used the F1-score, defined as the harmonic mean of precision and recall.  Many early bot-detection models had relatively high precision but low recall, inflating accuracy metrics.  With low recall, these models underestimate the scale of the bot infestation and disinformation problem in general. We found the F1 score as an adequate balance emphasizing both precision and recall.  F1 score for all models is provided in Table \ref{tab:major_test}.

From the results we see that the Botometer model has the highest precision of all comparison models, but lower recall and therefore lower relative F1 score.  The Debot algorithm was able to identify two of the bots, has perfect precision, but very low recall and F1 score.  The bot-hunter algorithms improve recall at a slight cost in precision, resulting in slightly higher F1 scores compared to other models.  

The benchmark classification datasets were balanced, while the bot training data was not. To account for this, random over-sampling of the data was performed during training. Graph-Hist was hand-tuned after the grid search used on the benchmark datasets, resulting in the final configuration given in Table \ref{tab:model_parameters}. The training environment was the same as that used for the benchmark dataset experiments. The new stopping threshold was given by F1 score in the validation set. Graph-Hist has recall higher than all other models and precision slightly below bot-hunter, resulting in the highest F1 score of all models tested.

\section{Conclusions}
In this paper, we have proposed a neural network architecture, Graph-Hist, for graph classification. Graph-Hist creates expressive node embeddings from GCNs in a similar manner to previous successful models, and uses a powerful CNN architecture to classify these embeddings in an end-to-end manner. While each aspect of the model has not appeared in its exact form in prior literature, the most significant innovation here is the binning module, which allows node embedding distributions to be approximated in a differential manner, such that convolutional architectures are then applicable. The binning procedure was inspired by the analysis of large social networks, and as such has been applied to social network classification tasks. Graph-Hist advances the state-of-the-art performance on 4 out of the 6 tested benchmark datasets, including all 3 of the social media benchmarks. 

Lastly, Graph-Hist was applied to a new graph classification domain: bot detection. 
Graph-Hist demonstrated better generalization in this task than the current leading bot detection models. 
Graph classification methods have another huge advantage to classic approaches when it comes to bot detection: they are hard to guard against.
These models are highly non-linear, so it is not obvious what types of graphs a ``puppet-master" should try to construct to avoid detection.
Even if an inconspicuous structure was known, the communication graph is far more challenging to manipulate than simple features like tweet frequency.
While communication networks are more costly to collect, the popularization of graph classification approaches to bot detection should slow down the ``cat and mouse" cycle we are currently experiencing. 

Future extensions of this work may involve attaching binning modules to different embeddings schemes, or classifying the resulting histograms with new methods. It could also include improvements in the bot domain, specifically by classifying other types of entities, such as trolls, which may have communication graphs differing from both normal actors and bots. More generally, future work could advocate for increased attention to social media datasets though the release of new social media benchmark datasets which reflect the scale and sparsity of networks seen in the wild. 

\section{ Acknowledgments}
This work was supported in part by the Office of Naval Research (ONR) Multidisciplinary University Research Initiative Award  N000140811186 and Award N000141812108, and the Center for Computational Analysis of Social and Organization Systems (CASOS). Thomas Magelinski was also supported by an ARCS Foundation scholarship. The views and conclusions contained in this document are those of the authors and should not be interpreted as representing the official policies, either expressed or implied, of the ONR, ARL, DTRA, or the U.S. government.

\bibliographystyle{aaai}
\bibliography{main}

\end{document}